\documentclass[journal,letterpaper]{IEEEtran}

\usepackage{cite}
\usepackage{amsmath,amssymb,amsfonts}
\usepackage[largesc]{newtxtext}
\usepackage{newtxmath}
\usepackage{algorithm}
\usepackage{algorithmic}
\usepackage{graphicx}
\usepackage{textcomp}
\usepackage{xcolor}
\usepackage{booktabs}
\usepackage{multirow}
\usepackage{array}
\usepackage[hidelinks]{hyperref}
\usepackage[capitalize]{cleveref}
\crefname{figure}{Fig.}{Figs.}
\Crefname{figure}{Fig.}{Figs.}
\usepackage{soul}

 

\newcommand{\model}{PAMoR}

\graphicspath{{figures/}}

\usepackage{cuted}
\usepackage{flushend}

\begin{document}

\title{\model{}: Parameterized Affective Motion \\ Generation in Real Time for Humanoid Robots}


\author{Yan~Pan, Lingfan~Bao, Tianhu~Peng and Chengxu~Zhou%
\thanks{This work was partially supported by the Advanced Research and Invention Agency (grant number SMRB-SE01-P06) and the NVIDIA Academic Grant Program.}
\thanks{The authors are with the Department of Computer Science, University
College London, London, UK. \tt\small
chengxu.zhou@ucl.ac.uk}%
}


\markboth{}{}
\pagestyle{empty}

\maketitle
\thispagestyle{empty}

\begin{strip}
  \vskip-3\baselineskip
  \centering
  \includegraphics[width=0.8\textwidth]{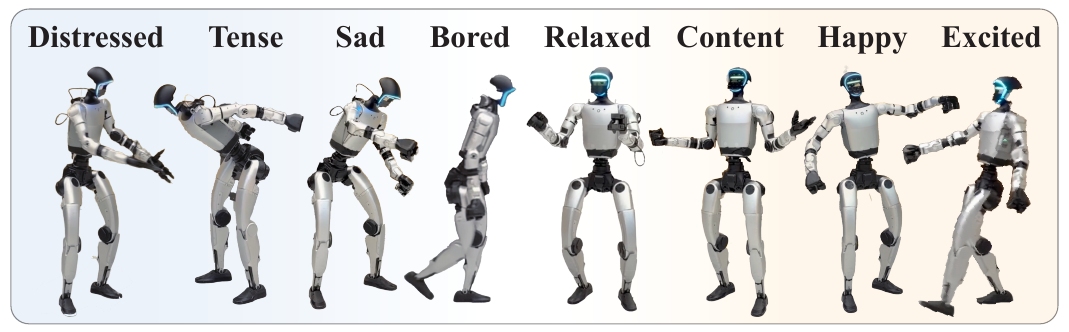}

  \refstepcounter{figure}\label{fig:teaser}
  {\footnotesize\textbf{Fig.~\thefigure.}\quad
  Affect as a continuous control parameter on the 29-DoF Unitree~G1.
  Each pose is a different commanded action and affect.}
\end{strip}

\begin{abstract}
People read a humanoid robot's motion in social settings not only for the action performed but for the affect conveyed. Motion carrying that affect has so far been generated for human avatars, where style is taken from a reference clip or an emotion word, neither of which can be quantitatively parameterized. We present \model{}, which turns affect into a measured control parameter: a valence--arousal (V-A) coordinate computed natively on robot kinematics. It is obtained in closed form from postural expansion and movement energy, and these measurements serve directly as generation conditions, with no human annotation. An action prior and two affect priors, trained in a shared latent space, are composed at each denoising step: the action prior fixes \emph{what} is performed, the affect priors modulate \emph{how}. Whole-body motion rolls out autoregressively on a 29-DoF Unitree~G1 in real time, with action and affect both editable. Generated motion tracks the commanded V-A over its full range while text-to-motion fidelity still matches text-only baselines. In a perceptual study, raters identify the commanded emotion on $0.38$ of trials, above both baselines and approaching the $0.44$ reported for acted human bodies.
\end{abstract}


\section{Introduction}
\label{sec:intro}

Beyond performing tasks, a robot is also a social agent whose movements people constantly read and react to~\cite{urakami2023}. As these robots enter everyday settings such as live entertainment and public-facing service~\cite{grandia2024bdx, yang2021hospitality}, motion becomes a channel that carries the robot's affect to whoever is watching~\cite{venture2019}. People assign meaning to that motion whether or not any was intended~\cite{urakami2023}. Controlling that affect takes three things. (a) The affect must be parameterized in a form that can be explained. (b) The motion must be produced in real time and stay editable while it runs. (c) The platform must be a humanoid, so that body-language knowledge still applies.

Work on generating affective body motion has concentrated on human avatars, where the emotion is specified by a reference clip~\cite{smoodi} or by an emotion word in the text prompt~\cite{emotionT2M}. With a reference clip, \emph{sad} is whatever that particular clip contains: it cannot be written down or reproduced, and the style it carries is entangled with the action it was performed on. With an emotion word, what \emph{sad} means is fixed by annotator consensus over the training corpus. The label has no definition at the level of motion. Either way, affect is not treated as an explainable, parameterized condition. Such a parameterization exists in psychology as the valence--arousal (V-A) plane~\cite{russell1980}. On robots, a few systems do command affect as a V-A value~\cite{moveae, sripathy2022}, but continuous V-A control for whole-body motion on legged humanoids remains underexplored. With no arms or torso to measure affect on, the scale behind the value can only come from hand-placed anchors or human annotation.

How the motion itself reaches a robot is a separate gap. Avatar-side generators produce a whole sequence offline, so the motion neither runs in real time nor changes once it is under way. Reaching the robot then requires retargeting, which adds latency in the inference loop and alters the postures and speeds that carry the affect. A correctly commanded emotion can be lost in the transfer. Generating natively on the humanoid eases the problems. Recent work already does so in real time, with the command editable during execution~\cite{xie2026textop}, but affect is not among its conditions.

\begin{figure*}
\centering
\includegraphics[width=\textwidth]{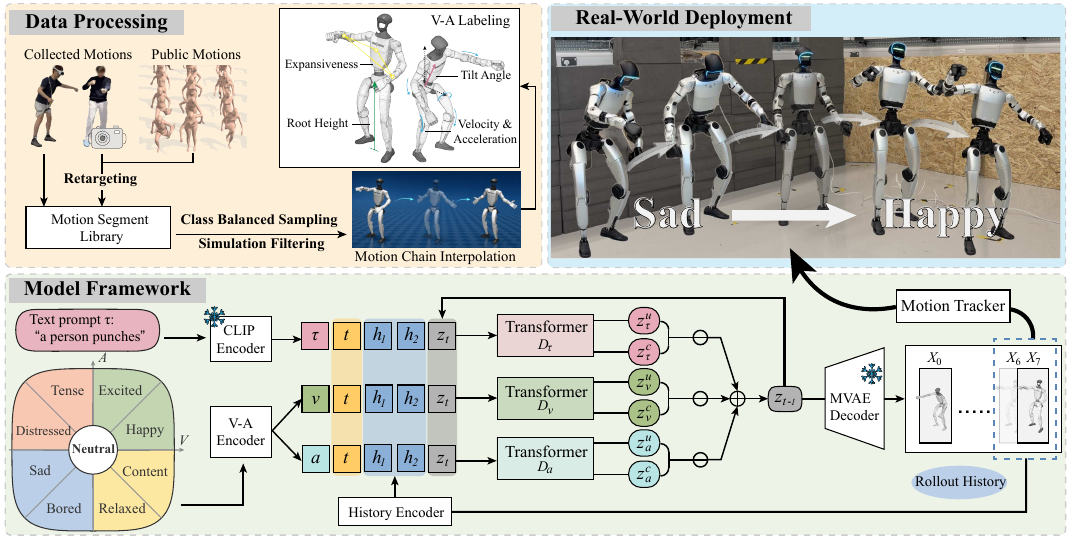}
\caption{Overview of our V-A-conditioned composable latent
diffusion framework.
\emph{Data Processing} (top left): public and video motions are
retargeted to the Unitree G1, teleoperated motions are recorded on it
directly, and all are stored as a motion-segment library;
class-balanced sampling, interpolation into longer chains, and simulation
filtering follow, and a continuous V-A label is computed in
closed form from body kinematics (arm expansiveness, root height, trunk
tilt angle, and velocity/acceleration). \emph{Model Framework} (bottom): a text prompt $\tau$ and a valence $v$ and arousal $a$ picked from the circumplex at left each condition a separate Transformer
denoising prior ($D_\tau,D_v,D_a$) that
shares the diffusion step $t$, motion history $h_1,h_2$, and noisy latent
$z_t$ in a frozen MVAE latent space. Their classifier-free guidance terms
are composed into a clean-latent prediction and noised back to $z_{t-1}$ for the next denoising step; the last step leaves the clean latent that the frozen MVAE decoder turns into the
next motion primitive, which is fed back as history for autoregressive rollout and executed on the robot by a motion tracker.
\emph{Real-World Deployment} (top right): the same action is performed
across the affect plane on a real G1, e.g., from \emph{Sad} to
\emph{Happy}.}
\label{fig:framework}
\end{figure*}

To close both gaps, we propose \model{}, a humanoid robot affective motion generation framework that works in two steps. First, affect is parameterized as a V-A coordinate measured directly from body kinematics. Valence measures how positive or negative the affect is, arousal how activated or calm~\cite{russell1980}. The first step applies the body-language findings normally used to recognize affect from motion~\cite{noroozi2021,lu2025}. An open and expanded posture reads as positive valence, and fast, energetic movement as high arousal~\cite{kleinsmith2013,karg2013}. We turn those findings into a closed-form function of the robot's forward kinematics that labels the entire training corpus with no human annotation. Second, the coordinate is turned into control. Conditioning a single network on text, valence, and arousal jointly entangles them, and adjusting one shifts the others, as our ablation confirms (\cref{sec:ablation}). We therefore take a composable approach~\cite{composable}. Three diffusion priors are trained separately in a shared motion latent space, each learning how a single condition shapes motion, and at inference their guidance is summed at each denoising step. The text prompt switches what is performed, while the valence and arousal values modulate how it is performed. Whole-body motion rolls out in real time on the 29-DoF Unitree G1, with all three conditions user-specified and changeable on the fly. An overview is shown in \cref{fig:framework}.

In summary, our contributions are:
\begin{itemize}
  \item A \emph{kinematics-grounded V-A parameterization} that bridges
  low-level body kinematics and high-level emotion through the continuous
  V-A plane, built on established body-language findings and requiring no
  human annotation.
  \item \emph{Real-time affective motion generation}
  on a humanoid, where three priors are composed at sampling time to combine
  mixed-type conditions, the text prompt for the action and two continuous
  affect scalars, so action and affect can be edited
  independently during the autoregressive rollout.
\item A \emph{real-robot user study} showing that the link between body
  movement and perceived emotion, established on humans, carries over to a
  humanoid robot: raters recognize the commanded emotions at rates approaching
  those reported for acted human bodies.
  \end{itemize}

\section{Related Work}
\label{sec:related}

\subsection{Motion Diffusion for Humanoid Robots}

Diffusion models cast text-to-motion as conditional denoising, and the approach matured first on human motion. Early models denoise in raw motion space~\cite{mdm}; later ones move into a learned latent space for efficiency and fidelity~\cite{mld}, and DART adds an autoregressive rollout that generates motion from text in real time~\cite{dart}. These models, however, generate in a human motion space, so each output must be retargeted to the robot~\cite{araujo2025gmr}. Retargeting in the inference loop adds latency~\cite{jia2026echo}. And for our task, a human-avatar motion generated at a commanded V-A would not match that value after retargeting, since the transfer alters the postures and speeds that carry affect. Recent humanoid systems therefore generate directly in the robot's own feature space and push retargeting offline into a one-time data-preparation step; ECHO and TextOp, for instance, synthesize motion in a robot-native representation and avoid retargeting at inference~\cite{jia2026echo,xie2026textop}. Our generator denoises in the robot's own latent space and rolls the motion out autoregressively, leaving execution to a whole-body tracker~\cite{sonic}. Unlike the systems above, it takes affect as a condition.

\subsection{Composable Diffusion}

The models above all condition on the text prompt alone. In our setting, action and affect must be controlled at once. A single network conditioned jointly on text, valence, and arousal entangles them. Modulating valence or arousal then distorts the action itself, as our ablation confirms (\cref{sec:ablation}). Composable diffusion avoids this by modeling each condition with a separate diffusion model and summing their scores at every denoising step, so the conditions are satisfied together while each stays separately controllable~\cite{composable}. The same composition has recently been applied to motion, for combining multiple textual concepts~\cite{energymogen}. Ours combines conditions of different kinds: one prior carries the action, two carry continuous affect scalars.

\subsection{Affective Motion Generation}

Affective and expressive motion has been generated in several ways, but conditioning whole-body humanoid generation on a continuous affect signal remains open. On human avatars, affect is specified by example or by label: SMooDi transfers the style of a reference clip onto motion generated from text~\cite{smoodi}, and emotion-enriched text-to-motion injects a named emotion at the limb level, turned into per-limb guidance by an LLM~\cite{emotionT2M}. In co-speech gesture, AMUSE drives emotional body motion from speech through latent diffusion~\cite{amuse}; it shapes a speaking-style gesture rather than the affect of a chosen action. In all of these, affect enters as an example or a label, not as an explainable, parameterized condition.

On robots, some work has explored parameterizing affect as valence and arousal. Sripathy et al.\ reach emotive behavior on the bipedal Cassie and a vacuum robot through offline trajectory optimization~\cite{sripathy2022}, and Suguitan et al.\ edit the valence and arousal of a fixed clip on the small robot Blossom by arithmetic in a latent space~\cite{moveae}. Marmpena et al.\ generate short emotional body expressions at a target V-A on the wheeled humanoid Pepper, using a valence-conditioned CVAE and setting arousal by latent-space sampling~\cite{marmpena2020generating}. Together, these works establish V-A-based robot motion control, but their platforms provide limited whole-body articulation compared with a full-size humanoid, and affect is applied by offline optimization, post-hoc editing, or short expressive animations rather than as a real-time generation condition on a chosen action. Closest in platform, HIAER runs on a humanoid and infers valence and arousal from the social scene with a vision-language model~\cite{hiaer}; there V-A drives intention inference and gesture selection rather than conditioning the generator. We make continuous V-A a direct condition of the generator itself and generate whole-body humanoid motion in real time.

\subsection{Grounding V-A in Body Kinematics}
\label{sec:va_ground}

Valence and arousal are the two axes of Russell's circumplex model of affect~\cite{russell1980}. Named emotions such as \emph{happy} or \emph{sad} occupy positions on this continuous plane rather than forming discrete categories. It is an empirical result that the body alone carries affect: Fourati and Pelachaud crossed seven daily actions with eight emotions and, with no facial cue available, observers still recovered the intended emotion well above chance~\cite{emilya}, a protocol our rater study adapts to the robot (\cref{sec:user_study}). Which movement kinematics carry which axis is also established rather than assumed. Across surveys of bodily affect the same regularity recurs: valence shows in posture, an open, expanded body for positive states and a contracted, collapsed one for negative~\cite{kleinsmith2013,karg2013}, while arousal follows movement energy, correlating with velocity and acceleration~\cite{karg2013}, and is reported as the easier of the two to read from the body~\cite{kleinsmith2013}. Building on this link, prior work trains recognizers that estimate affect from body motion~\cite{noroozi2021,lu2025}. This line, however, targets recognition rather than generation, and it learns from subjective human annotations. We instead compute V-A directly from the grounded indicators in closed form over the robot's own kinematics, giving each training motion an interpretable, continuous V-A label without manual affect annotation (\cref{sec:va_embed}).

\section{Method}
\label{sec:method}

In this section, we present our framework for V-A-conditioned motion generation, illustrated in \cref{fig:framework}.

\subsection{Problem Formulation}
\label{sec:problem}

We represent motion in the robot's own feature space, where each frame is a feature vector $\mathbf{x}_t$ stacking root orientation, heading change, foot contacts, heading-invariant root displacement, root height, joint angles, and joint deltas:
\begin{equation}
\mathbf{x}_t \;=\;
\bigl[\,\mathbf{r}_t^{\sin\cos},\; \Delta\psi_t,\; \mathbf{f}_t,\;
       \Delta\mathbf{b}_t^{xyz},\; b_t^{z},\; \mathbf{q}_t,\; \Delta\mathbf{q}_t \,\bigr]
\in \mathbb{R}^{69}.
\label{eq:feature}
\end{equation}
Given a text prompt $\tau$ specifying an action, such as \emph{walk} or \emph{wave arms}, a target V-A coordinate $(v,a)\!\in\![-1,1]^2$, and a motion history $\mathbf{h}\!=\!\mathbf{x}_{t-H:t-1}$ of $H$ past frames, the goal is to synthesize the next $F$-frame primitive $\mathbf{x}_{t:t+F-1}$ for the $29$-DoF Unitree G1 and to roll it out autoregressively into a full motion. The action is determined by $\tau$, while $(v,a)$ modulates the emotional style.

\subsection{V-A Labeling}
\label{sec:va_embed}

\begin{figure}[t]
\centering
\includegraphics[width=\columnwidth]{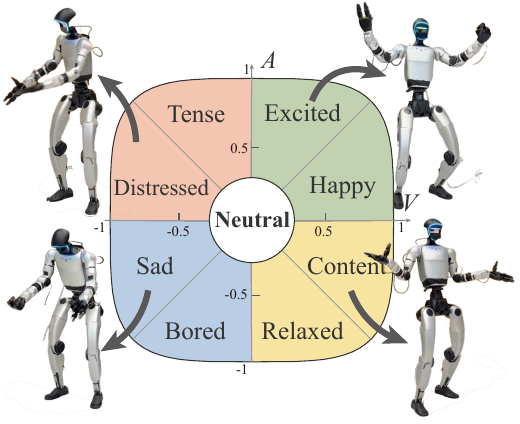}
\caption{The V-A plane realized on the robot. Valence runs left to
right, arousal bottom to top, with Russell's eight emotions at their
circumplex positions~\cite{russell1980}, neutral at the origin, and one generated motion per
quadrant. Valence sets how
contracted or open the body is, arousal how still or dynamic the motion.}
\label{fig:va_plane}
\end{figure}

Valence and arousal are computed rather than annotated. Both are evaluated in closed form on the $14$ body keypoints obtained from $\mathbf{x}_t$ by forward kinematics, taking posture for valence (how positive the expressed affect is) and movement energy for arousal (how activated it is), as illustrated in \cref{fig:va_plane}. This labels the entire training corpus at no annotation cost and keeps the V-A condition interpretable.

Valence is built from the postural expansion established in \cref{sec:va_ground}, using the indicators that literature identifies~\cite{kleinsmith2013,karg2013} and reading them off the robot's own kinematics. We measure the expansion with three cues, one per direction in which the body opens. Arm expansiveness $s_t$ is the perimeter of the triangle formed by the two wrists and a torso keypoint, and captures the lateral opening. Root height $b_t^{z}$ captures the vertical extension. Trunk tilt angle $\theta_t$, the signed pitch of the root in its yaw-aligned frame, captures the sagittal opening. It is positive when the trunk leans back and grows more negative the further the body folds forward. All three rise as the body opens. The literature grounds the dimension but gives no quantitative weights for the cues, so we weight them equally. Each cue is $z$-normalized over all training frames (written $\tilde{\cdot}$) and the three are summed,
\begin{equation}
v_t^{\text{raw}} \;=\; \tilde{s}_t \;+\; \tilde{b}_t^{z} \;+\; \tilde{\theta}_t.
\label{eq:v_raw}
\end{equation}
Arousal is built the same way from movement energy~\cite{karg2013}, again on the same keypoints. Speed $\dot{p}_t$ averages the velocity magnitudes of all keypoints, and acceleration $\ddot{p}_t$ their acceleration magnitudes. Both are $z$-normalized as above and summed,
\begin{equation}
a_t^{\text{raw}} \;=\; \widetilde{\dot{p}}_t \;+\; \widetilde{\ddot{p}}_t.
\label{eq:a_raw}
\end{equation}
Both raw scores are unbounded, so a winsorized rescaling maps each onto $[-1,1]$, anchored at the $10$th and $90$th percentiles $q_{10}$ and $q_{90}$ of that score over all training frames,
\begin{equation}
v_t \;=\; \operatorname{clip}\!\left(
   2\,\frac{v_t^{\text{raw}} - q_{10}}{q_{90} - q_{10}} - 1,\; -1,\; 1
\right),
\label{eq:winsor}
\end{equation}
and $a_t$ likewise. The per-frame labels are averaged over each $F$-frame primitive to give the $(v,a)$ conditions.

\subsection{Composable Latent Diffusion}
\label{sec:latent_diff}

Our generator denoises in a motion latent space and composes three conditional priors at each step (\cref{fig:framework}).

\subsubsection{Autoencoder}
We do not run diffusion on the raw features $\mathbf{x}_t$. A $9$-layer Transformer Motion VAE (MVAE)~\cite{mld} compresses each window of $H$ history and $F$ future frames into a $128$-dimensional latent token $\mathbf{z}$. The decoder reconstructs the $F$ future frames from the token and the history frames.

\subsubsection{Composable Conditional Priors}
A single prior conditioned jointly on text, valence, and arousal entangles action with affective style, so changing the V-A condition can also distort the requested action. We therefore train three separate priors that share the same noisy latent $\mathbf{z}_t$, diffusion timestep $t$, and motion history $\mathbf{h}$, but use different conditions:
\begin{equation}
D_\tau(\mathbf{z}_t; t, \tau, \mathbf{h}),\;\;
D_v(\mathbf{z}_t; t, v, \mathbf{h}),\;\;
D_a(\mathbf{z}_t; t, a, \mathbf{h}).
\label{eq:priors}
\end{equation}
From here on $t$ is the diffusion timestep rather than a frame index: $\mathbf{z}_t$ is the clean MVAE latent $\mathbf{z}_0$ noised to level $t$, and denoising runs from $t=T$, pure noise, down to $t=0$. Each prior predicts the clean latent directly rather than the noise added to it~\cite{mdm}. The text prior learns the action specified by $\tau$, whereas the affect priors learn how the V-A conditions change motion style.

Each affect prior is trained on pairs whose condition is the scalar measured from that primitive's own motion, so what it fits is the distribution of clean latents at a given value of that scalar. Since $v$ measures postural expansion, high $v$ selects the latents that decode to an open, extended body and low $v$ those that decode to a folded one; $a$, measuring movement energy, likewise separates fast motion from slow. At inference the difference between a prior's conditioned and unconditioned prediction~\cite{ho2022cfg} is a direction in latent space that pulls its scalar toward the commanded value. Because the three priors are fit separately, the valence direction is estimated without reference to the action, so following it changes how the motion is performed while the text prior holds what is performed (\cref{sec:experiments}).

The three priors use the same Transformer denoiser architecture and differ only in their condition and their independently trained weights. Each denoiser reads a single token sequence $[\,\mathbf{c},\,t,\,\mathbf{h}_1,\,\mathbf{h}_2,\,\mathbf{z}_t\,]$. The sequence holds one conditioning token, the diffusion timestep, one token per history frame, and the noisy latent. The denoiser attends over the whole sequence and predicts the clean latent. Conditions thus enter in context as tokens rather than through cross-attention. A frozen CLIP text encoder~\cite{clip} embeds the action prompt $\tau$. Linear layers map valence, arousal, and the two history frames to their tokens.

At each denoising step we combine them via classifier-free composition~\cite{ho2022cfg},
\begin{equation}
\hat{\mathbf{z}}_0
   \;=\; \hat{\mathbf{z}}_\tau^{u}
   \;+\; \sum_{i\,\in\,\{\tau,v,a\}} \lambda_i\,
         \bigl(\hat{\mathbf{z}}_i^{c} - \hat{\mathbf{z}}_i^{u}\bigr),
\label{eq:compose}
\end{equation}
where $\hat{\mathbf{z}}_0$ is the composed clean-latent prediction, prior $i$ predicts $\hat{\mathbf{z}}_i^c$ with its condition and $\hat{\mathbf{z}}_i^u$ without it, and $\lambda_i$ scales its guidance term. This is the composition rule of composable diffusion~\cite{composable}, which assumes the three conditions are independent given the motion. Every denoising step forms $\hat{\mathbf{z}}_0$ this way and noises it back to $\mathbf{z}_{t-1}$; repeating from $t=T$ leaves the clean latent $\mathbf{z}_0$, which the decoder maps to a motion primitive. Each primitive is sampled under its own conditions and its last $H$ frames become the history for the next, so a rollout runs to any length while the prompt and the V-A target may change between primitives (\cref{fig:framework}).

\subsubsection{Training Objectives}
We train the model in two stages. A single MVAE is trained first, under a reconstruction term over the motion representation together with forward-kinematic and temporal-difference terms, and a KL regularizer on the latent. That one MVAE is then frozen and encodes the latents for all three priors, so their predictions all live in the same space and can be summed. On those shared latents the three priors are trained independently of one another, each with a diffusion loss that regresses the clean latent,
\begin{equation}
\mathcal{L}_i
   \;=\; \mathbb{E}\bigl[\,
         \mathcal{H}\!\bigl( \mathbf{z}_0,\; D_i(\mathbf{z}_t;\, t, c_i, \mathbf{h}) \bigr)
         \,\bigr],
   \qquad i \in \{\tau, v, a\},
\label{eq:diff_loss}
\end{equation}
where the expectation runs over training latents, timesteps, and noise draws, $c_i$ is that prior's condition, and $\mathcal{H}$ is the Huber loss. Alongside it the same forward-kinematic and temporal-difference terms are applied to the motion decoded from the predicted latent. During training we drop the condition with probability $0.1$, which supplies $\hat{\mathbf{z}}_i^u$. The priors compose only at sampling time, so one can be added or removed without retraining.

\subsection{Motion Corpus Construction}
\label{sec:corpus}

The V-A priors need examples of the same action performed in different kinematic styles. BABEL-annotated AMASS motions~\cite{amass,babel} cover many actions, but not evenly in style. Frequent classes contain many variants, whereas rare ones span only a narrow range of speed, amplitude, and posture. We therefore add motions reconstructed from video by GVHMR~\cite{gvhmr} and teleoperated motions recorded on the robot. Both target everyday expressive actions such as spread arms and handshake, performed in several styles each. The video and teleoperated motions were performed by members of the research team, who consented to the use of their motion data.

The AMASS and video sources are retargeted to the Unitree G1 with GMR~\cite{araujo2025gmr}, whereas teleoperated motions are already in G1 space (\cref{fig:framework}). All sources are cut into single-action segments, which we group into $15$ action classes. Three segments are then interpolated into one chain, with classes drawn at equal rates so that frequent actions do not crowd out rare ones. Chaining exposes the model to transitions between actions, which single-action segments alone would not provide, and each chain keeps the action label and text of every segment it contains. Segments and chains alike are replayed in simulation with the SONIC tracker~\cite{sonic}, and whatever it fails to track, or that leaves the robot below a minimum root height or past a maximum lean, is discarded.
\section{Experiments}
\label{sec:experiments}

We first compare motion quality and text alignment against baselines with standard text-to-motion metrics~\cite{humanml3d}. We then evaluate V-A controllability, measuring whether generated motions match the commanded affect. We further run a user study on perceived emotion. Finally, we ablate the composable and latent designs.

\subsection{Experimental Setup}
\label{sec:datasets}

\subsubsection{Dataset}
The corpus of \cref{sec:corpus} combines BABEL-annotated AMASS motions~\cite{amass,babel} with video-reconstructed and teleoperated recordings. It comprises $10{,}095$ training and $2{,}406$ held-out chains over the $15$ action classes, all on the Unitree G1 at $20$\,Hz. Each chain interpolates three segments, giving $30{,}285$ training and $7{,}218$ held-out segments.

\subsubsection{Evaluator and metrics}
We compute all metrics in a shared text--motion embedding space using a G1-TMR evaluator, adapted from TMR~\cite{Petrovich2023tmr} and trained on robot-space G1 motions. Following the standard text-to-motion protocol~\cite{humanml3d}, we report FID, R-precision, MM-Distance, Diversity, and MultiModality; Diversity measures variation across different actions, whereas MultiModality measures variation across generations of the same action. Action accuracy, reported in the ablation, scores whether the action itself survives: each generation is assigned to the nearest action-class text by the same evaluator, and we report the fraction assigned to the requested class. These metrics use $1{,}000$ annotated segments drawn at random from the held-out split under a fixed seed, so every method is scored on the same prompts, with $30$ generations per action class for MultiModality. The baselines are conditioned on text alone; ours additionally receives a valence and an arousal value drawn uniformly from $[-0.8,0.8]$.

For V-A controllability we sweep valence and arousal each over seven values evenly spaced across the same range and generate at all $49$ combinations for every action prompt, three seeds apiece. For the user study we report three measures against the emotion each clip was commanded to convey (\cref{sec:user_study}). Top-3 is the fraction of trials on which that emotion appears among the rater's ranked three. $\kappa_w$ scores the first pick alone. With $n_{ij}$ the trials commanded at $i$ and answered $j$, $e_{ij}$ the same count expected if answers ignored the command, and $d_{ij}$ the steps from $i$ to $j$ around the circumplex,
\begin{equation}
\kappa_w = 1 - \frac{\sum_{ij} d_{ij}^{2}\, n_{ij}}{\sum_{ij} d_{ij}^{2}\, e_{ij}},
\label{eq:kappa_w}
\end{equation}
which is $1$ at perfect agreement and $0$ at chance. The squared weight is what separates a near miss from a real one: calling \emph{excited} ``happy'' is one step and costs a ninth of calling it ``sad'', which is three. Naturalness is the mean five-point rating, from robotic to human-like.

\subsubsection{Implementation}
Every network is a Transformer with hidden size $512$. The MVAE carries about $42$M parameters and each of the three priors is an $8$-layer denoiser of about $17.5$M, trained over a $100$-step DDPM schedule~\cite{ddpm}. At inference we take $10$ of those steps per primitive, roll out $F\!=\!8$ frames at a time from $H\!=\!2$ history frames, and compose the three priors with guidance weights $(\lambda_\tau,\lambda_v,\lambda_a)\!=\!(5,2,2)$.

\subsection{Quantitative Evaluation}
\label{sec:quantitative}

\subsubsection{Baseline comparison}

A text-only model produces the most typical version of the requested action, which is exactly what the evaluator matches best to the text, so these metrics flatter it. A V-A target pushes generations away from that typical version and should therefore cost FID and R-precision. It does not: our FID, R-precision, MM-Distance and Diversity stay on par with the text-only baselines while MultiModality rises (\cref{tab:main}), so V-A conditioning adds style variation at no measurable cost to motion quality.

We compare against two robot-space motion diffusion baselines, both conditioned only on text: TextOp~\cite{xie2026textop}, an autoregressive latent-diffusion model with a Transformer denoiser, and ECHO~\cite{jia2026echo}, a motion-space diffusion model with a 1D-convolutional U-Net denoiser. We also compare against SMooDi~\cite{smoodi}, a stylized motion diffusion model.

\begin{table}[t]
\centering
\caption{Text-to-motion evaluation. All methods are evaluated on one fixed
set of $1{,}000$ held-out prompts. For SMooDi only R-precision is comparable.}
\label{tab:main}
\footnotesize
\setlength{\tabcolsep}{2.8pt}
\renewcommand{\arraystretch}{1.15}
\begin{tabular}{lcccccc}
\toprule
Method & FID $\scriptstyle\downarrow$ & R@1 $\scriptstyle\uparrow$ & R@3 $\scriptstyle\uparrow$ & MM-Dist $\scriptstyle\downarrow$ & Div. $\scriptstyle\rightarrow$ & MMod. $\scriptstyle\uparrow$ \\
\midrule
Real (ceiling)                & $0.000$ & $0.985$ & $0.992$ & $1.106$ & $1.305$ & $1.094$ \\
\midrule
TextOp~\cite{xie2026textop}   & $0.438$ & $0.799$ & $0.904$ & $1.188$ & $1.240$ & $0.521$ \\
ECHO~\cite{jia2026echo}       & $0.366$ & $0.798$ & $0.897$ & $1.187$ & $1.272$ & $0.787$ \\
SMooDi~\cite{smoodi}          & --      & $0.166$ & $0.309$ & --      & --      & --      \\
Ours                          & $0.379$ & $0.792$ & $0.883$ & $1.188$ & $1.274$ & $\mathbf{0.882}$ \\
\bottomrule
\end{tabular}
\end{table}

\subsubsection{An affect-conditioned baseline}
\label{sec:smoodi}

To our knowledge, no action-conditioned affective motion generator has been released for a legged whole-body humanoid, so the closest available system is SMooDi~\cite{smoodi}, which applies the style of a reference clip to content given by a text prompt. We retarget its human motion to the G1 as in \cref{sec:corpus}. Its style clips come from the 100STYLE library~\cite{style100}, which is not our corpus and matches only four of our eight circumplex wedges. The comparison therefore uses those four, and scores SMooDi only on R-precision, since the distribution-based metrics depend on the training data.

Style comes at the cost of the action: R@1 falls to $0.166$, against roughly $0.79$ for the three methods trained on our corpus (\cref{tab:main}). SMooDi's own ablation reports the same exchange: enabling its style guidance lifts style accuracy from $0.202$ to $0.724$ and drops R@3 from $0.630$ to $0.571$~\cite{smoodi}.

\subsubsection{V-A control}
We command a V-A value and measure it back on the generated motion with the labeling of \cref{sec:va_embed}. On both axes the measured value tracks the commanded one, at a rank correlation of $0.95$. The slope is $+1.10$ for valence and $+0.79$ for arousal. The response is monotone over the whole range, so any point of the circumplex~\cite{russell1980} can be commanded rather than a few labelled emotions. Cross-axis correlations stay below $0.06$, so commanding one axis does not drift the other.

\subsubsection{Inference efficiency}
On a single RTX~5090, the full three-prior rollout takes $78$\,ms per primitive at $T\!=\!10$ denoising steps, and each primitive is $400$\,ms of motion.

\begin{figure}[t]
\centering
\includegraphics[width=\columnwidth]{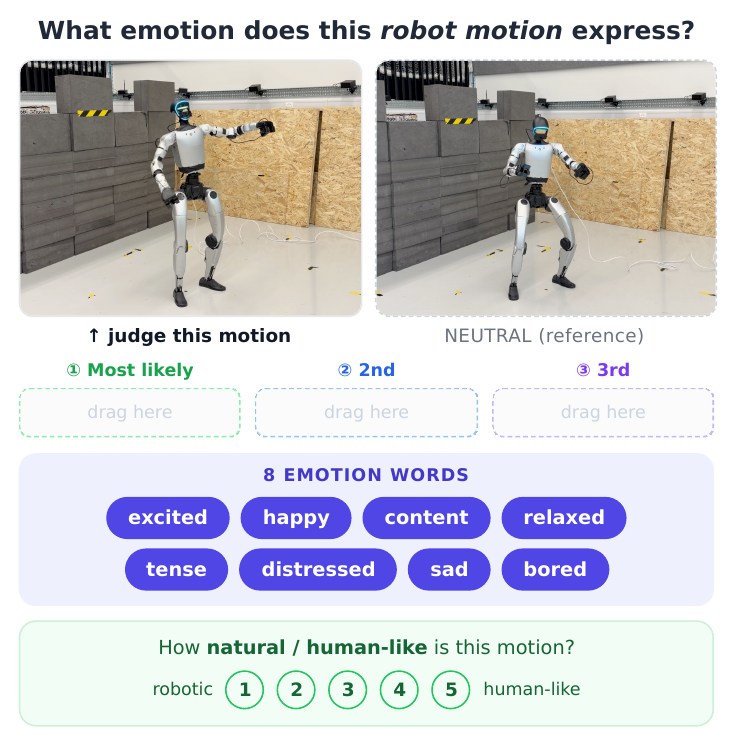}
\caption{User-study interface. The clip to be judged sits left of a neutral reference of the same action; the rater drags their top three emotion guesses into ranked slots from eight options and rates naturalness on a five-point scale.}
\label{fig:user_study_ui}
\end{figure}

\subsection{User Study}
\label{sec:user_study}

We run a perceptual study in which raters identify the emotion conveyed by video of the G1 and rate its naturalness. This study constitutes minimal-risk product and user testing under the research ethics policy of the authors' institution and therefore did not require ethical review. Participants viewed only pre-recorded videos of the robot and provided non-sensitive perceptual ratings. All participants provided informed consent.

\subsubsection{Setup}
We use seven actions: handshake, spread arms, wave arms, punch, walk, clap, and kick. The eight emotions are Russell's own terms~\cite{russell1980}, one per $45^\circ$ wedge of the circumplex: happy, excited, tense, distressed, sad, bored, relaxed, and content. We compare three ways of asking a generator for an emotion: TextOp with an emotive prompt asks in words, as in ``punch happily''; SMooDi asks by example, through a reference clip; and ours asks by value, through a point on the circumplex.
The protocol follows the one Fourati and Pelachaud established for perceiving affect in the body~\cite{emilya}: actions are fully crossed with emotions, so affect must be read from \emph{how} an action is performed rather than from which action it is, and the response set is the full set of eight emotions on every trial, whatever the stimulus, so chance stays at $1/8$ for the first pick and $3/8$ for the ranked three. As in their study no facial cue is available, in our case by construction, since the G1 has no face.
Each trial shows two clips of the same action side by side, the emotional clip to be judged and a neutral reference, so what is rated is the affect added on top of the action (\cref{fig:user_study_ui}). Raters rank the three most likely emotions out of the eight and score naturalness on a $5$-point scale. Ours and the emotive prompt cover all eight emotions on all seven actions, at $56$ clips each, and SMooDi covers the four wedges its style library reaches (happy, excited, tense, and sad), at $28$, for $140$ clips in all. 

The $140$ clips were split into two sets of $70$, each holding half the clips from every method, balanced by action and as evenly as possible by emotion. Twelve raters, unaffiliated with the project and naive to the hypotheses and to the methods compared, were recruited in person at the university and assigned six to each set. All completed their $70$ assigned trials, with no exclusions, giving six independent judgements per clip and $840$ trials in total. Trials from the three methods were shown in random order, and the method behind each clip was never identified. Raters gave informed consent and responses were collected anonymously; under the authors' institutional research ethics policy the study was assessed as minimal-risk research not requiring ethical review (details withheld for anonymous review).

\begin{table}[t]
\centering
\caption{Perceptual study. Chance is $0.125$ for Top-1 and $0.375$ for Top-3.}
\label{tab:user_study}
\footnotesize
\setlength{\tabcolsep}{4pt}
\renewcommand{\arraystretch}{1.1}
\begin{tabular}{lcccc}
\toprule
Method & Top-1 $\uparrow$ & Top-3 $\uparrow$ & $\kappa_w$ $\uparrow$ & Natural. $\uparrow$ \\
\midrule
TextOp~\cite{xie2026textop}, emotive prompt & $0.125$ & $0.423$ & $0.060$ & $3.40$ \\
SMooDi~\cite{smoodi}               & $0.202$ & $0.417$ & $0.209$ & $3.25$ \\
Ours                               & $\mathbf{0.384}$ & $\mathbf{0.845}$ & $\mathbf{0.688}$ & $\mathbf{4.05}$ \\
\bottomrule
\end{tabular}
\end{table}

\subsubsection{Results}
Every trial ranks three of the eight words, so a rater who ignored the clip
would still carry the commanded emotion into their three on $3/8$ of them, and
name it outright on $1/8$. Ours is named first on $0.384$ of trials and carried
into the three on $0.845$, both clear of chance on a $95\%$ interval, at $0.33$
to $0.44$ and $0.80$ to $0.88$ (\cref{tab:user_study}). Neither baseline
separates from chance on the ranked three: both land a little above $3/8$ but
inside their own intervals. The emotive prompt does not separate on any
measure, and only SMooDi's first pick clears a line, at $0.202$. Scoring the
first pick with chance removed and near misses weighted, $\kappa_w$ separates
the three the same way, at $0.688$ for ours against $0.209$ and $0.060$. The ordering also holds inside
every rater: all twelve ranked ours above the emotive prompt on Top-3, so the
gap does not rest on a subset of observers. Where ours is
misread, the word chosen is a neighbour on the circumplex rather than an
arbitrary one, at a mean of $1.48$ wedges from the command against the $2.29$
expected under random answering; the baselines miss at $2.20$ and $2.15$,
essentially the random figure.
The affect is thus read in the right direction and missed only in
degree, which is a distinction a continuous command can express and a discrete
label cannot. The two baselines fall short for different reasons. The emotive
prompt has no affect to draw on, because the text in our corpus names the
action and never the manner. SMooDi does carry affect, but its style clips
reach only four of the eight wedges, and transferring style costs the action
itself (\cref{sec:smoodi}). Naturalness stays near $3$ of $5$ or above for all
three, so conditioning on V-A does not make the motion read as less human-like.

For comparison, acted human emotional body
expressions in the Emilya database are recognized by observers on $0.44$
of stimuli against the same $1/8$ chance line, with per-emotion recall spread
from $0.12$ for shame to $0.87$ for sadness~\cite{emilya}. Motion
generated for the G1 is therefore read at rates approaching those of
acted human bodies, over a per-emotion range of $0.21$ for relaxed to
$0.49$ for excited. Recognition varies by action
in both studies over a comparable band, $0.29$ to $0.46$ here against $0.36$ to
$0.48$ there.

One difference from that setting is worth noting. Emilya's actions are
affect-neutral by construction, and even so the action shapes what is read from
it: anger is recovered on $1.00$ of throwing and $0.89$ of lifting trials but
only $0.25$ of walking ones, which the authors attribute to the action affording
the emotion~\cite{emilya}. Two of our actions instead work against the command,
since punching and kicking are read as aggressive before any affect is added.
Commanding them from the positive half of the plane is therefore the harder
case, and the command still comes through: where we commanded positive, the
raters' first pick was positive on $0.85$ of trials, while over all punching
and kicking trials their first picks split $49$ positive to $47$ negative,
with no lean toward either half.

\begin{table}[t]
\centering
\caption{Single-prior versus composable diffusion under V-A modulation.}
\label{tab:ablation_joint}
\footnotesize
\setlength{\tabcolsep}{5pt}
\renewcommand{\arraystretch}{1.1}
\begin{tabular}{lccc}
\toprule
Variant & FID $\downarrow$ & R@1 $\uparrow$ & Action Acc. $\uparrow$ \\
\midrule
Single prior $(\tau,v,a)$             & $0.447$ & $0.556$ & $0.478$ \\
\textbf{Composable priors (ours)}  & $\mathbf{0.379}$ & $\mathbf{0.792}$ & $\mathbf{0.748}$ \\
\bottomrule
\end{tabular}
\end{table}

\subsection{Ablation Study}
\label{sec:ablation}

\subsubsection{Composable versus single-prior diffusion}
Conditioning a single prior on text and V-A jointly entangles affective control with action identity. Under V-A modulation the single-prior variant loses more than $0.2$ on both text retrieval and action accuracy (\cref{tab:ablation_joint}), whereas our composable priors preserve the requested action while V-A modulates style.

\subsubsection{Latent versus motion-space diffusion}
The motion-space variant keeps the same composable design but denoises raw features instead of latents. Both are scored on the V-A sweep of \cref{sec:datasets}. The slopes $s_v$ and $s_a$ relate the measured V-A to the commanded value and should sit at $1$. The on-axis correlations $\rho_{v\to v}$ and $\rho_{a\to a}$ rise as an axis follows its own command more tightly, and joint jerk falls as the motion smooths. Denoising in the frozen MVAE latent makes the V-A command both more visible and smoother. Slopes sit closer to $1$, on-axis correlations rise on both axes (arousal $0.79\!\to\!0.95$), and joint jerk drops by $41\%$ (\cref{tab:ablation_latent}).

\begin{table}[t]
\centering
\caption{Latent versus motion-space diffusion under the same composable
design. Jerk is in rad/s$^3$.}
\label{tab:ablation_latent}
\scriptsize
\setlength{\tabcolsep}{2.6pt}
\renewcommand{\arraystretch}{1.1}
\begin{tabular}{lccccc}
\toprule
Variant & $s_v{\to}1$ & $s_a{\to}1$ & $\rho_{v\to v}\uparrow$ & $\rho_{a\to a}\uparrow$ & Jerk $\downarrow$ \\
\midrule
Motion-space  & $+1.122$ & $+0.570$ & $0.896$ & $0.786$ & $157.8$ \\
\textbf{Latent (ours)} & $\mathbf{+1.095}$ & $\mathbf{+0.787}$ & $\mathbf{0.950}$ & $\mathbf{0.947}$ & $\mathbf{92.7}$ \\
\bottomrule
\end{tabular}
\end{table}
\section{Discussion}
\label{sec:discussion}

The grounding literature fixes which cues raise valence and which raise arousal, but not how much each should count. The equal weighting in~(\ref{eq:v_raw}) is therefore a convention, and calibrating it will require further experiments.

At deployment the framework is hierarchical: the model generates the motion and the whole-body controller executes it, with nothing fed back. The controller inevitably tracks with some error and has no notion of affect, so the executed V-A can drift from the command with nothing in place to detect it. Whether closing this loop holds the commanded affect on the real robot is the question we would take up first.

The system already generates in real time and keeps action and affect editable while the motion runs, which is the foundation interactive use needs. Future work is to apply it where the value comes from reasoning about the scene~\cite{hiaer} and adjusts to the person's feedback.

\section{Conclusion}
\label{sec:conclusion}

We presented \model{}, which conditions real-time humanoid motion generation on valence and arousal measured in closed form from the robot's own kinematics. Generated motions follow the commanded V-A coordinates across the evaluated range and are executed on the physical Unitree G1. The requested action is largely preserved under V-A modulation, and raters recognize the intended emotions more reliably than with the evaluated baselines. The ablation indicates that the proposed composable design better preserves action identity than the jointly conditioned model, which exhibits substantially lower action accuracy. These results indicate that parameterizing affect from the body itself, rather than through annotation, is what makes the condition explainable and verifiable.


\bibliographystyle{IEEEtran}
\bibliography{references}

\end{document}